\documentclass[sigconf,nonacm=true]{acmart}
\renewcommand\footnotetextcopyrightpermission[1]{} % removes footnote with conference information in first column
\AtBeginDocument{%
  \providecommand\BibTeX{{%
    \normalfont B\kern-0.5em{\scshape i\kern-0.25em b}\kern-0.8em\TeX}}}

\setcopyright{acmcopyright}
\copyrightyear{2019}
\acmYear{2018}
\acmDOI{10.1145/1122445.1122456}

\usepackage{textcomp}
\usepackage{multirow}

\begin{document}

%%
%% The "title" command has an optional parameter,
%% allowing the author to define a "short title" to be used in page headers.
\title{GateNet:Gating-Enhanced Deep Network for Click-Through Rate Prediction}

%%
%% The "author" command and its associated commands are used to define
%% the authors and their affiliations.
%% Of note is the shared affiliation of the first two authors, and the
%% "authornote" and "authornotemark" commands
%% used to denote shared contribution to the research.
% \author{Tongwen Huang}
% \authornote{Work done during at Sina Weibo.} 
% \affiliation{%
%   \institution{Sina Weibo Corp}
%   \city{Beijing}
%   \country{China}
% }
% \email{tongwenzide@163.com}

\author{Tongwen Huang}
\authornote{Work done at Sina Weibo.} 
\affiliation{
     \institution{Tencent Corp, China}}
\email{towanhuang@tencent.com}

\author{Qingyun She, Zhiqiang Wang, Junlin Zhang}
\affiliation{
     \institution{Sina Weibo Corp, China}}
\email{{qingyun,zhiqiang36,junlin6}@staff.weibo.com}

%%
%% By default, the full list of authors will be used in the page
%% headers. Often, this list is too long, and will overlap
%% other information printed in the page headers. This command allows
%% the author to define a more concise list
%% of authors' names for this purpose.
\renewcommand{\shortauthors}{Tongwen Huang et al.}

%%
%% The abstract is a short summary of the work to be presented in the
%% article.
\begin{abstract}
Advertising and feed ranking are essential to many Internet companies such as Facebook. Among many real-world advertising and feed ranking systems, click through rate (CTR) prediction plays a central role. In recent years, many neural network based  CTR models have been proposed and achieved success such as Factorization-Machine Supported Neural Networks, DeepFM and xDeepFM. Many of them contain two commonly used  components: embedding layer and MLP hidden layers. On the other side, gating mechanism is also widely applied in many research fields such as computer vision(CV) and natural language processing(NLP). Some research has proved that gating mechanism improves the trainability of non-convex deep neural networks. Inspired by these observations, we propose a novel model named GateNet which introduces either the feature embedding gate or the hidden gate to the embedding layer or hidden layers of DNN CTR models, respectively. The feature embedding gate provides a learnable feature gating module to select salient latent information from the feature-level. The hidden gate helps the model to implicitly capture the high-order interaction more effectively. Extensive experiments conducted on three real-world datasets demonstrate its effectiveness to boost the performance of various state-of-the-art models such as FM, DeepFM and xDeepFM on all datasets.
\end{abstract}

\maketitle

\section{Introduction}
Advertising and feed ranking are essential to many Internet companies such as Facebook. 
The main technique behind these tasks is click-through rate prediction which is known as CTR. 
Many models have been proposed in this field such as logistic regression (LR)\cite{mcmahan2013ad}, 
polynomial-2 (Poly2)\cite{juan2016field}, tree based models\cite{he2014practical}, tensor-based models\cite{koren2009matrix}, 
Bayesian models\cite{graepel2010web}, and factorization machines based models\cite{rendle2010factorization,juan2016field}.

With the great success of deep learning in many research fields such as
computer vision\cite{krizhevsky2012imagenet} and natural language processing\cite{mikolov2010recurrent,cho2014learning}, many
deep learning based CTR models have been proposed in recent
years\cite{zhang2016deep,cheng2016wide,guo2017deepfm,lian2018xdeepfm,zhang2019fat}. % (TODO:add the gate introduction)
Many of them contain two commonly used components:embedding layer and MLP hidden layers. On the other side, gating mechanism is also widely applied in many research fields such as computer vision(CV) and natural language processing(NLP). Some research works have proved that  gating  mechanism  improves  the  trainability of non-convex deep neural networks. Inspired by these observations, a model named GateNet is proposed to select salient latent information from the feature-level and implicitly capture the high-order interaction more effectively for CTR prediction. 

Our main contributions are listed as follows:
\begin{itemize}
\item We propose the feature embedding gate layer to replace the traditional embedding and enhance the model ability. 
Inserting the feature embedding gate into the embedding layer of many classical models such as FM, DeepFM, DNN and XDeepFM, 
we observe a significant performance improvement.
\item The MLP layers are an essential component to implicitly capturing the high-order feature interaction in the canonical DNN models, 
we introduce the hidden gate to the MLP parts of deep models and improve the performance of the the classical models.
\item It is simple and effective to enhance the standard DNN model by inserting hidden gate
and we can achieve comparable performance with other state-of-the-art model baselines such as DeepFM and XDeepFM.
\end{itemize}

The rest of this paper is organized as follows. In Section \ref{sec:s2}, we review
related works which are relevant with our proposed model, followed by
introducing our proposed model in Section \ref{sec:s3}. We will present
experimental explorations on three real-world datasets in Section \ref{sec:s4}. Finally, we conclude this work in Section \ref{sec:s5}.

\section{Related Work}
\label{sec:s2}
\subsection{Deep Learning based CTR Models}
Many deep learning based CTR models have also been proposed in recent years\cite{zhang2016deep,cheng2016wide,guo2017deepfm,lian2018xdeepfm,wang2017deep}. How to effectively model the feature interactions is the key factor for most of these neural network based models. Factorization-Machine Supported Neural Networks (FNN)\cite{zhang2016deep} is a forward neural network using FM to pre-train the embedding layer.
However, FNN can capture only high-order feature interactions. Wide \& Deep model(WDL)\cite{cheng2016wide} jointly trains wide linear models and deep neural networks to combine the benefits of memorization and generalization for recommendation systems. 
However, expertise feature engineering is still needed on the input to the wide part of WDL. 
To alleviate manual efforts in feature engineering, DeepFM\cite{guo2017deepfm} replaces the wide part of WDL with FM and shares the
feature embedding between the FM and deep component.

In addition, Deep \& Cross Network (DCN)\cite{wang2017deep} and eXtreme Deep Factorization Machine (xDeepFM)\cite{lian2018xdeepfm} are recent deep learning methods which explicitly model the feature interactions.

\begin{figure*}[hbt!]
\centering
\includegraphics[width=15.36cm, height=5.98cm]{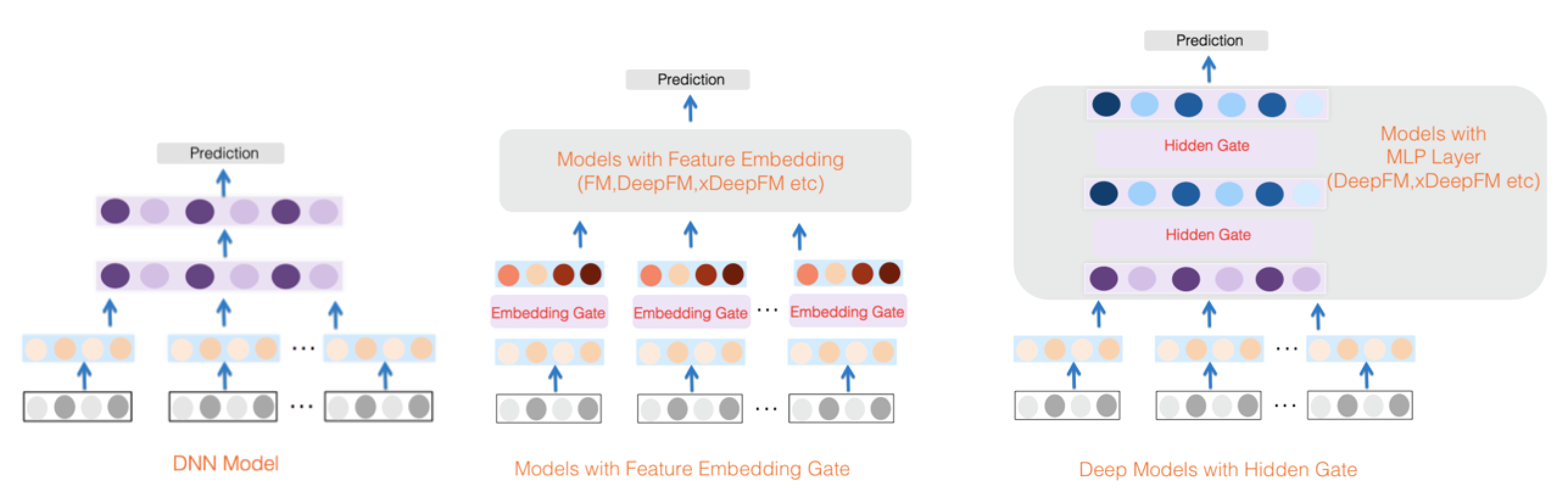}

\caption{The architecture of our proposed GateNet. The left diagram is the standard DNN network, the middle of diagram is the models with feature embedding gate and the right diagram is the deep models with hidden gate.}
\label{fig:f1}
\end{figure*}

\subsection{Gating Mechanisms in Deep Learning}
Gating mechanism is widely used in many deep learning fields, such as computer vision(CV), natural language processing(NLP), and recommendation systems.

The gate mechanism is used in computer vision, such as Highway Network \cite{srivastava2015highway}, they utilize the transform gate and the carry gate to express how much of the output is produced by
transforming the input and carrying the output, respectively.

The gate mechanism is widely applied to NLP, such as LSTM\cite{gers1999learning}, GRU\cite{cho2014learning}, language modeling\cite{dauphin2017language}, sequence to sequence learning\cite{gehring2017convolutional}
and they utilize the gate to prevent the gradients vanishing and resolve the long-term dependency problem.

In addition, \cite{ma2018modeling} uses the gates to automatically adjust parameters between modeling shared information and modeling task-specific information in recommendation systems. Another recommendation system applying the gate mechanism is hierarchical gating network(HGN)\cite{ma2019hierarchical} and they apply feature-level and instance-level gating
modules to adaptively control what item latent features and which relevant item can be passed to the downstream layers.

\section{Our Proposed Model}
\label{sec:s3}

Deep learning models are widely used in industrial recommendation systems, such as WDL, YouTubeNet\cite{covington2016deep} and DeepFM.
The DNN model is a sub-component in many current DNN ranking systems, and its network structure is shown in the left of Figure \ref{fig:f1}.

% As mentioned above,
We can find two commonly used components in most of the current DNN ranking systems: the embedding layer and MLP hidden layer. 
% (TODO:to explain the functions or motivations of feature embedding gate and hidden gate)
We aim to enhance the model ability and propose the model named GateNet for CTR prediction tasks. 
First, we propose the feature embedding gating layer which can convert embedding features into gate-aware embedding features and helps to select salient latent information from the feature-level. 
Second, we also propose the hidden gate which can adaptively control what latent features and which relevant feature interaction can be passed to the downstream layer.
The DNN model with feature embedding gate and DNN model with hidden gate are depicted as the middle and right in Figure \ref{fig:f1}. In the following subsections, we will describe the feature embedding layer and hidden gate layer in GateNet in detail.

\subsection{Feature Embedding Gate}
\label{sec:s31}
The sparse input layer and embedding layer are widely used in deep learning based CTR models such as DeepFM\cite{guo2017deepfm}.
The sparse input layer adopts a sparse representation for raw input features. 
The embedding layer is able to embed the sparse feature into a low dimensional, dense real-value vector. 
The output of embedding layer is a wide concatenated field embedding vector:
$$E=[e_1, e_2, \cdots,e_i, \cdots, e_f]$$
where $f$ denotes the number of fields, $e_i \in R^k$ denotes the embedding of $i$-th field, and $k$ is the dimension of embedding layer.

On the other side, recent research results show that gate can improve the train-ability in training non-convex deep
neural networks\cite{glorot2010understanding}.
% Inspired by these ideas, 
In this work, firstly we propose the feature embedding gate to select salient latent information from the feature-level in the DeepCTR model. 
The basic steps of the feature embedding gate can be described as followed:

First, for every field embedding $e_i$, we calculate the gate value which represents the feature-level importance of embedding. 
We formalize this step as the following formula: 
\begin{equation}
  g_i = \sigma(W_{i} \cdot e_i)
\end{equation}
where $\sigma$ is the activation function of gate, $e_i \in R^{k}$ is the original embedding, $W_{i}$ is the learned parameters of the $i$-th gate and the total number of learned parameter matrix $W=[W_1, \cdots,W_i, \cdots, W_f]$, $i=1, \cdots, f$.

Second, we assign the gate value to the corresponding feature embedding and generate a gate-aware embedding.
\begin{equation}
  ge_i = e_i \odot g_i 
\end{equation}
where $\odot$ denotes the Hadamard or element-wise product, $e_i \in R^{k}$ is the $i$-th original embedding, $i=1, \cdots, f$.

Third, we collect all gate-aware embeddings and regard it as gated feature embedding.
\begin{equation}
  GE=[ge_1, ge_2, \cdots,ge_i, \cdots, ge_f]
\end{equation}

\begin{figure}[hbt!]
   \includegraphics[width=8cm, height=5cm]{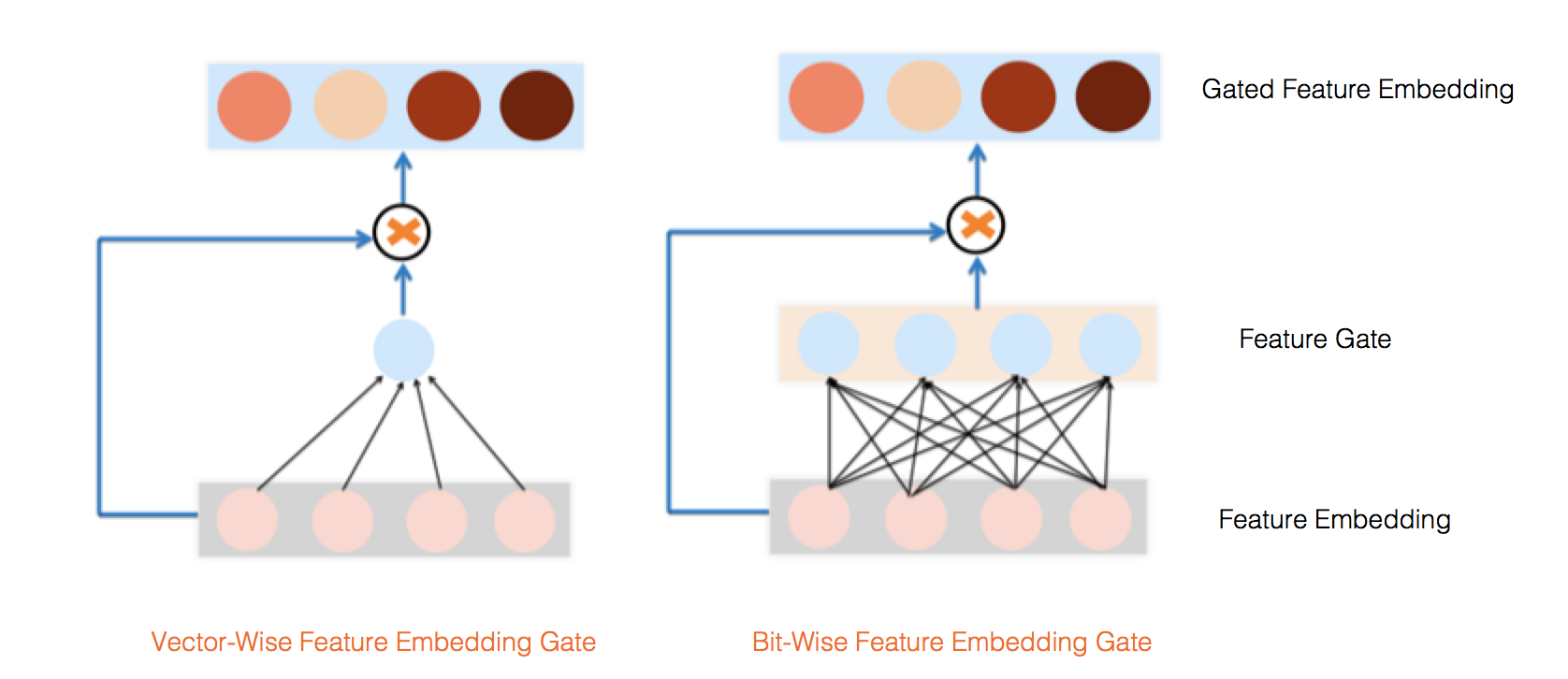}
  \caption{The Feature Embedding Gate. The left diagram represents vector-wise feature embedding gate and the right diagram is bit-wise feature embedding gate.}
  \label{fig:e1}
\end{figure}

It is a common practice to make gate output a scalar which represents the importance of the whole feature embedding.
To learn the bit level salient important information in the feature embedding, we can make this gate output a vector
which contains fine-grained information about the feature embedding.
And we call this embedding gate `bit-wise' gate and the common gate `vector-wise' gate. 
The vector-wise and bit-wise feature embedding gate can be depicted  as Figure \ref{fig:e1}.

Seen from the figure, we compare the difference of vector-wise feature gate and bit-wise feature is as follows:
\\
Vector-wise: $g_i \in R$, $W_{i} \in R^{k \times 1}$, $W \in R^{f \times k \times 1}.$ \\
Bit-wise: $g_i \in R^{k}$, $W_{i} \in R^{k \times k}$, $W \in R^{f \times k \times k}.$\\

We can see that the output of bit-wise gate is a vector which is related to each bit of feature embedding and it can be regarded as using the same value to each bit of feature embedding. The performance comparison of vector-wise and bit-wise feature embedding gate will be discussed in Section \ref{sec:s42}.

Moreover, as some previous works such as FiBiNet\cite{DBLP:conf/recsys/HuangZZ19} does, we will explore the parameter sharing mechanism of the feature embedding gate layer. 
Each gate in the feature embedding gate layer has its own parameters to explicitly learn the salient feature information, 
we also can make all the gates share parameters in order to reduce the number of parameters. 
We call this gate  `field sharing' and previous gate `field private'.
From a mathematical perspective, the biggest difference between `field sharing' and `field private' is the learned gate parameters $W_i$. 
$W_i$ is shared among all the fields in `field sharing' while $W_i$ is different for each field in `field private'.
The performance of `field sharing' and `field private' will be compared in Section \ref{sec:s42}.

\subsection{Hidden Gate}
\label{sec:s35}
The deep part of many DNN ranking systems usually consists of several full-connected layers, which implicitly captures high-order features interactions. 
As shown in Figure \ref{fig:f1}, the input of deep network is the flatten of embedding layer. 
Let $a^{(0)} = [ge_1,\cdots, ge_i, \cdots, ge_f]$ denotes the outputs of embedding layer, where $ge_i \in R^k$ represents the $i-$th feature embedding. 
Then, $a^{(0)}$ is fed into multi-layer perceptron network, and the feed forward process is:
\begin{equation}
    a^{(l)}=\sigma(W^{(l)}a^{(l-1)}+b^{(l)})
\end{equation}
where $l$ is the depth and $\sigma$ is the activation function. $W^{(l)}$,$b^{(l)}$,$a^{(l)}$ are the model weight, bias
and output of the $l$-th layer. 

\begin{figure}[hbt!]
  % \centering
  \includegraphics[width=7cm, height=4.4cm]{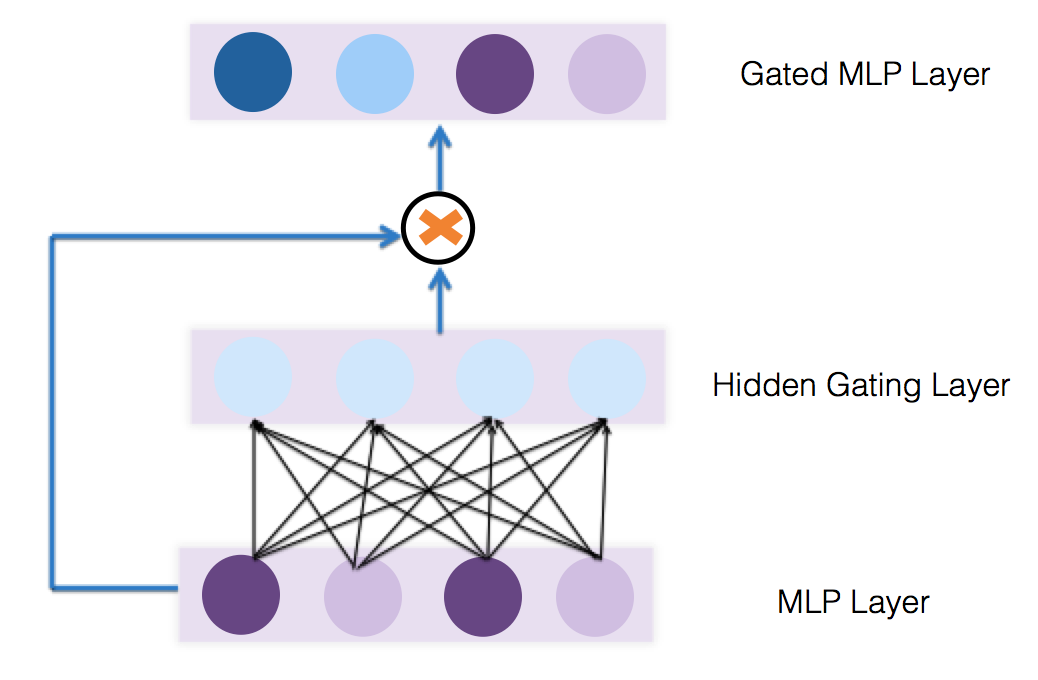}
  \caption{The Hidden Gate Layer}
  \label{fig:d1}
\end{figure}

Similar to the bit-wise feature embedding gate, we proposed the hidden gate which can be applied to the hidden layer. As depicted as Figure \ref{fig:d1}, we use this gate as follows:
\begin{equation}
  g^{(l)}=a^{(l)} \odot \sigma_g(W_g^{(l)}a^{(l)})
\end{equation}
where $\odot$ denotes the element-wise product, $\sigma_g$ is the gate activation function, $W_g^{(l)}$ is the $l$-th layer parameter of hidden gate. Likewise, we can stack multiple hidden gate layers like the classic DNN models.

\subsection{Output Layer}
To summarize, we give the overall formulation of our proposed model'output as:
\begin{equation}
    \hat{y}=\sigma(W^{|L|}g^{|L|} + b^{|L|})
\end{equation}
where $\hat{y} \in(0,1)$ is the predicted value of CTR, $\sigma$ is the sigmoid function, $b^{|L|}$ is the bias and $|L|$ is the depth of DNN. The learning process aims to minimize the following objective function
(cross entropy):
\begin{equation}
   loss = {-}\frac{1}{N}\sum_{i=1}^{N}({y_i}log(\hat{y_i})+(1-y_i)*log(1-\hat{y_i}))
\end{equation}
where $y_i$ is the ground truth of $i$-th instance, $\hat{y_i}$ is
the predicted CTR, and $N$ is the total size of samples.

\section{Experiments}
\label{sec:s4}
In this section, we conduct extensive experiments to answer the following research questions:

\noindent\textbf{(RQ1)} Can the feature embedding gate enhance the ability of the baseline models?

\noindent\textbf{(RQ2)} Can the hidden gate enhance the ability of the baseline models?

\noindent\textbf{(RQ3)} Can we combine the two gates in one model to achieve further improvements?

\noindent\textbf{(RQ4)} How do the settings of networks influence the performance of our model?

We will answer these questions after presenting some fundamental experimental settings.
\subsection{Experimental Testbeds and Setup}
\subsubsection{Data Sets} 1) Criteo.
The Criteo\footnote{http://labs.criteo.com/downloads/} dataset is widely used in many CTR model evaluation. 
It contains click logs with 45 millions data instances. 
There are 26 anonymous categorical fields and 13 continuous feature fields in Criteo dataset.
We split the dataset randomly into two parts: 90\% is for training,
while the rest is for testing. 2) ICME. The ICME\footnote{https://biendata.com/competition/icmechallenge2019}
dataset consists of several days of short video click datas. It contains click logs with 19 millions data instances in track2. 
For each click data, we choose 5 fields(user\_id, user\_city, item\_id,author\_id,item\_city) to predict the like probability of short video.
We split it randomly into two parts: 70\% is for training, while the rest is for testing. 3) SafeDriver. The SafeDriver\footnote{https://www.kaggle.com/c/porto-seguro-safe-driver-prediction}
dataset is used to predict the probability that an auto insurance policy holder files a claim.
There are 57 anonymous fields in SafeDriver dataset and these features are divided into similar groups:binary features, categorical features, continuous features and ordinal features. 
It contains 595K data instances. We split the dataset randomly into two parts: 90\% is for training, while the rest is for testing.
\subsubsection{Evaluation Metrics}
In our experiment, we adopt AUC(Area Under ROC) as metric. AUC is a widely used metric in evaluating classification problems. Besides, some work validates AUC as a good measurement in CTR prediction\cite{graepel2010web}. 
AUC is insensitive to the classification threshold and the positive ratio. 
The upper bound of AUC is 1, and the larger the better. 

\subsubsection{Baseline Methods}
To verify the effect of the gate layer added in various mainstream models, we choose some widely used CTR models as our baseline models including FM\cite{rendle2010factorization,rendle2012factorization}, DNN, DeepFM\cite{guo2017deepfm}, and XDeepFM\cite{lian2018xdeepfm}.

Main goal of this work is not intent to propose a new model instead of enhancing these baseline models via gating mechanism that we proposed.
\noindent Note that an improvement of 1\textperthousand\ in AUC is usually regarded as
significant for the CTR prediction because it will bring a large
increase in a company's revenue if the company has a very large user base.

\subsubsection{Implementation Details}
\label{sec:414}
We implement all the models with Tensorflow\footnote{TensorFlow:
  https://www.tensorflow.org/} in our experiments. For the embedding
layer, the dimension of embedding layer is set to 10. For the optimization method, we use the Adam\cite{kingma2014adam} with a mini-batch size of 1000, and the learning rate is set to 0.0001. For all deep models, the depth of layers is set to 3, all activation functions are RELU, the number of neurons per layer
is 400, and the dropout rate is set to 0.5.
The default activation function of feature embedding gate is Sigmoid and activation function of hidden gate is Tanh.    
We conduct our experiments with 2 Tesla K40 GPUs.

\subsection{Performance of Feature Embedding Gate(RQ1)}
\label{sec:s42}
In this subsection, we show the performance gains of chosen baseline models after inserting feature embedding gate into a typical embedding layer. The experiments are conducted on Criteo,ICME and SafeDriver datasets and results are shown in Table \ref{table:t1}.
  
\begin{table}[ht]
  \centering
  \caption{The overall performance improvements of baseline models after inserting feature embedding gate into a typical embedding layer. Unless specially mentioned in our paper, `field private' and `vec-wise' model are used as the default setting in the embedding gate model. The suffix with `$e$' stands for applying the embedding gate to this model.}
  \label{table:t1}
  \begin{tabular}{llll}
  Model & ICME & Criteo & SafeDriver \\ \hline
  FM & 0.8696 & 0.7923 & 0.6302 \\
  FM$_e$ & 0.8973 & 0.7970 & 0.6327 \\
  $\Delta$ & 0.0277 & 0.0047 & 0.0025 \\ \hline
  DNN & 0.8912 & 0.8067 & 0.6344 \\
  DNN$_e$  & 0.9166 & 0.8096 & 0.6359 \\
  $\Delta$ & 0.0254 & 0.0029 & 0.0015 \\ \hline
  DeepFM & 0.9027 & 0.8087 & 0.6276 \\
  DeepFM$_e$  & 0.9097 & 0.8097 & 0.6349 \\
  $\Delta$ & 0.0070 & 0.0010 & 0.0073 \\ \hline
  XDeepFM & 0.9052 & 0.8091 & 0.6324 \\
  XDeepFM$_e$  & 0.9178 & 0.8098 & 0.6336 \\
  $\Delta$ & 0.0126 & 0.0007 & 0.0012 \tabularnewline \bottomrule
  \end{tabular}
\end{table}

Inserting the feature embedding gate into these baseline models, we find our proposed embedding gate mechanisms can  consistently boost the baseline model's performance on these three datasets as shown in Table \ref{table:t1}. 
These results indicate that carefully selecting salient latent information from the feature-level is useful to enhance the model ability and make the baseline models achieve better performance.
Among all the baseline models, FM with the feature embedding gate gets a significant improvement which outperforms the classic FM model by almost 2\% on ICME dataset.
We assume that FM is a shallow model that has only a set of latent vectors to learn, 
there's no other component in FM to explicitly or implicitly adjust the feature in FM, 
so the gate mechanism is a good way to adjust the feature weight. 
Instead of FM, there are many deep models such as DeepFM and XDeepFM, our models with feature embedding gate can enhance these models' ability and make further improvements.

Moreover, we design some further research about feature embedding gate. 
First, we conduct some experiments to compare parameter sharing mechanism of gate(`field sharing' and `field private') in Table \ref{table:t2}.
\begin{table}[!htbp]
  \caption{Parameter sharing mechanism of feature embedding gate: field private vs field sharing.}
  \label{table:t2}
  \centering
  \begin{tabular}{lcccr}
    \toprule
     &\multicolumn{2}{c}{ICME}&\multicolumn{2}{c}{Criteo}\\
    \hline
     Model & Private & Share & Private & Share \\
    \midrule
FM & 0.8973 & 0.8861 & 0.7970 & 0.7957\tabularnewline
DNN & 0.9166 & 0.9076 & 0.8096 & 0.8099\tabularnewline
DeepFM & 0.9097 & 0.8985 & 0.8097 & 0.8098\tabularnewline
XDeepFM & 0.9178 & 0.9039 & 0.8098 & 0.8096\tabularnewline
  \bottomrule
\end{tabular}
\end{table}

From the Table \ref{table:t2}, we can find that the performance of `field private' gate is much better than the `field sharing' gate for many base models on ICME dataset while it is not significant on Criteo dataset.
Although the `field sharing' can reduce the number of learned parameters, the performance also decreases. 
These results indicate that the performance of different parameter sharing mechanisms of gate depend on specific task. 
On the whole, it is a better choice to choose the `field private' in our experiments. 

Second, we conduct some experiments to explore the vector-wise and bit-wise feature embedding gate.
\begin{table}[!htbp]
  \caption{Embedding gate mechanism: vector-wise vs bit-wise. The default gate share mechanism is `field private'.}
  \label{table:t3}
  \centering
  \begin{tabular}{lcccr}
    \toprule
     &\multicolumn{2}{c}{ICME}&\multicolumn{2}{c}{Criteo}\\
    \hline
     Model & vec-wise & bit-wise & vec-wise & bit-wise \\
    \midrule
FM & 0.8973 & 0.8937 & 0.7970 & 0.7985\tabularnewline
DNN & 0.9166 & 0.9018 & 0.8096 & 0.8098\tabularnewline
DeepFM & 0.9097 & 0.9112 & 0.8097 & 0.8098\tabularnewline
XDeepFM & 0.9178 & 0.9175 & 0.8098 & 0.8100\tabularnewline
  \bottomrule
\end{tabular}
\end{table}
The results in Table \ref{table:t3} show that bit-wise is a little better than vector-wise on Criteo dataset, while we cannot draw an obvious conclusion on the ICME data. 
The reason behind this needs further exploration.

\subsection{Performance of Hidden Gate(RQ2)}
In this subsection, the overall performance gains of chosen baseline models after inserting hidden gate into a typical MLP layer will be reported on these three test sets in Table \ref{table:t4}. 

% In this subsection, we summarize the 
\begin{table}[!htbp]
  \centering
  \caption{The overall performance improvements of baseline models with hidden gates. The suffix with `$h$' stands for applying the hidden gate to this model.}
  \label{table:t4}
  \begin{tabular}{llll}
  Model & ICME & Criteo & SafeDriver \\ \hline
  DNN & 0.8912 & 0.8067 & 0.6344 \\
  DNN$_h$  & 0.9105 & 0.8093 & 0.6348 \\
  $\Delta$ & 0.0193 & 0.0026 & 0.0004 \\ \hline
  DeepFM & 0.9027 & 0.8087 & 0.6276 \\
  DeepFM$_h$  & 0.9121 & 0.8090 & 0.6324 \\
  $\Delta$ & 0.0094 & 0.0003 & 0.0048 \\ \hline
  XDeepFM & 0.9052 & 0.8091 & 0.6324 \\
  XDeepFM$_h$  & 0.9084 & 0.8092 & 0.6344 \\
  $\Delta$ & 0.0032 & 0.0001 & 0.0020 \tabularnewline \bottomrule 
  \end{tabular}
\end{table}

Replacing the traditional MLP with the hidden gate layer, our proposed hidden gate mechanisms
consistently enhance these baseline models and achieve performance improvements on the ICME, Criteo and SafeDriver dataset as shown in Table \ref{table:t4}. The experimental results indicate that the hidden gate helps the model to implicitly capture the high-order interaction more effectively.

Although applying hidden gate to MLP layers is simple, it is an effective way to improve the performance of baseline models. 
Therefore, we conduct experiments to compare hidden gate DNN with some complex base models in the Table \ref{table:t41}.

\begin{table}[!htbp]
\centering
\caption{Compare the performance between standard DNN by inserting hidden gate and other base models. The Safe denotes SafeDrive dataset and XDFM represents XDeepFM.}
\label{table:t41}
\begin{tabular}{llllll}
Dataset  & DNN & DeepFM & XDFM & FiBiNet & DNN$_{h}$ \\ \hline
Criteo & 0.8063 & 0.8087 & 0.8091 & 0.8102 & 0.8093 \\
ICME  & 0.8912 & 0.9027 & 0.9052 & 0.9030 & 0.9105 \\
Safe  & 0.6344 & 0.6276 & 0.6324 & 0.6342 & 0.6348 \tabularnewline
  \bottomrule
\end{tabular}
\end{table}

From the Table \ref{table:t41}, the standard DNN by inserting hidden gate outperforms some canonical deep learning models such as DeepFM, XDeepFM.
It is a simple way to enhance the standard DNN to gain improvement, 
which makes the DNN model much more practicable in industrial recommendation systems.

\subsection{Performance of model Combining FE-Gate and Hidden Gate(RQ3)}

As mentioned previously, we find the feature embedding gate and hidden gate can enhance the model ability and gain good performance, respectively.
Can we combine the feature embedding gate and hidden gate in one model to achieve further performance?
We conduct some experiments to answer this research question on Criteo and ICME datasets. 
  
 \begin{table}[!htbp]
  \centering
  \caption{The overall performance of baseline models with feature embedding gate and hidden gate models.The `EGate', `HGate', `Both' denote the feature embedding gate, hidden gate and feature embedding gate plus hidden gate, respectively.}
  \label{table:t5}
\begin{tabular}{llllll}
Dataset & Model & Base & EGate & HGate & Both \\ \hline
\multirow{3}{*}{ICME}
 & DNN & 0.8912 & 0.9166 & 0.9195 & 0.9054\\
 & DeepFM & 0.9027 & 0.9097 & 0.9121 & 0.9114  \\
 & XDeepFM & 0.9052 & 0.9178 & 0.9084 & 0.9054 \\ \hline
\multirow{3}{*}{Criteo} 
 & DNN & 0.8067 & 0.8096 & 0.8093 & 0.8097 \\
 & DeepFM & 0.8087 & 0.8097 & 0.8090 & 0.8097 \\
 & XDeepFM & 0.8091 & 0.8098 & 0.8092 & 0.8098 \tabularnewline \bottomrule
\end{tabular}
\end{table}

It can be seen from Table \ref{table:t5} that combining feature embedding gate and hidden gate in one model can not 
gain further performance improvements. Specifically, there is not much performance improvements on Criteo and some performance decrease on ICME.
The feature embedding gate can influence the implicit and explicit feature interaction while the hidden gate can influence the implicit feature interaction,
we assume that the implicit feature interactions have been done twice and the implicit feature representations are damaged.
The real reason behind this need to conduct further experiments to justify this assumption.

\subsection{Hyper-parameter Study(RQ4)}
We conduct some experiments to study the influence of hyper-parameter in our proposed gate mechanisms.
We test different settings in our proposed GateNet on the SafeDriver dataset and 
we treat DeepFM, DeepFM$_e$ and DeepFM$_h$ as the baseline models.

So we divide the hyper-parameters into the following three parts: 
\begin{itemize}
  \item Gate activation function. Both embedding and hidden gate include the gate activation functions.
  \item Embedding size. We change the embedding size from 10 to 50, and compare the performance of baseline model with embedding gate model.
  \item Hidden layers. We change the number of layers from 2 to 6, and observe the performance of baseline model and hidden gate model.
\end{itemize}

\subsubsection{Activation function in Gate}
\begin{table}[!htbp]
  \centering
  \caption{The overall performance of different activation functions in the feature embedding gate and hidden gate.}
  \label{table:t6}
  \begin{tabular}{llllll}
    Model & Linear & Relu & Sigmoid & Tanh\\ \hline
    DeepFM$_{e}$ & 0.6356 & 0.6343 & 0.6320 & 0.6349  \\
    DeepFM$_{h}$ & 0.6321 & 0.6320 & 0.6311 & 0.6324  \tabularnewline \bottomrule
\end{tabular}
\end{table}
The test results on SafeDriver dataset with different activation functions in the feature embedding gate and hidden gate are presented in Table \ref{table:t6}.
We observe that the best activation function in feature embedding gate is linear while the best activation function is Tanh in hidden gate.

\subsubsection{Embedding Size in Feature Embedding Gate}

\begin{table}[!ht]
\centering
\caption{The performance of different embedding sizes in Feature Embedding gate.}
\label{table:t7}
\begin{tabular}{llllll}
Model & 10 & 20 & 30 & 40 & 50 \\ \hline
DeepFM & 0.6276 & 0.6297 & 0.6271 & 0.6284 & 0.6235 \\
DeepFM$_e$ & 0.6349 & 0.6322 & 0.6319 & 0.6329 & 0.6307 \tabularnewline
\bottomrule
\end{tabular}
\end{table}

We change the embedding size from 10 to 50 in feature embedding gate and summarize the range of performances in Table \ref{table:t7}. % and Figure \ref{fig:ex1}.
From the results, we find that embedding size has little influence on the GateNet.
Specifically, the standard DeepFM has a good performance with the embedding size 20, while the embedding size of DeepFM$_e$ is 10.
Therefore, these results show that DeepFM$_e$ requires less parameter than DeepFM to train a good model.

\subsubsection{Number of Layers in Hidden Gate}

In deep part, we can change the number of neurons per layer, depths of DNN, activation functions and dropout rates.
For brevity, we just study the impact of different depths in DNN part.
We change the number of layers from 2 to 6 in hidden gate and conclude the performance in Table \ref{table:t8}.% and Figure \ref{fig:ex2}.

\begin{table}[!ht]
\centering
\caption{The performance of different number of layers in DNN.}
\label{table:t8}
\begin{tabular}{llllll}
\#Layers & 2 & 3 & 4 & 5 & 6 \\ \hline
DeepFM & 0.6219 & 0.6276 & 0.6281 & 0.6290 & 0.6279 \\
DeepFM$_h$ & 0.6328 & 0.6324 & 0.6312 & 0.6321 & 0.6286 \tabularnewline
\bottomrule
\end{tabular}
\end{table}

Increasing the number of layers, the performance of DeepFM increases, while DeepFM$_h$ decreases.
These results indicate that our DeepFM$_h$ can learn much better than DeepFM with less parameters on SafeDriver dataset.

\section{Conclusions}
\label{sec:s5}
Recently, many neural network based CTR models have been proposed and some recent research results found that gating mechanisms can improve the trainability in training non-convex deep neural networks. Inspired by these observations, we proposed a novel model named GateNet which introduces either the feature embedding gate or the hidden gate to the embedding layer or hidden layers of DNN CTR models,respectively. Extensive experiments conducted on three real-world datasets demonstrate its effectiveness to boost the performance of various state-of-the-art models such as FM, DeepFM and xDeepFM on three real-world datasets.

\bibliographystyle{ACM-Reference-Format}
\bibliography{sample-base}
\end{document}